\documentclass[12pt,journal,draftcls,letterpaper,onecolumn]{IEEEtran}
\usepackage{graphicx}

\begin{document}

%opening
\title{Detecting lateral genetic material transfer}
\author{C. Calder\'on$^1$, L. Delaye$^2$, V. Mireles$^{3}$ and P. Miramontes$^{1,4,*}${\footnote{Corresponding author. e-mail:\texttt{ pmv@ciencias.unam.mx}}}\\
1. Facultad de Ciencias, Universidad Nacional Aut\'onoma de M\'exico, M\'exico 04510 DF, M\'exico\\
2. Departamento de Ingenier\'ia Gen\'etica, Centro de Investigaci\'on y Estudios Avanzados, Irapuato, Guanajuato, 36821, M\'exico.\\
3. International Max Planck Research School for Computational Biology and Scientific Computing. Freie Universit\"at Berlin, D-14195 Berlin, Germany\\
4. Centro de Ciencias de la Complejidad, Universidad Nacional Aut\'onoma de M\'exico, M\'exico 04510 DF, M\'exico}

\maketitle

\begin{abstract}

The bioinformatical methods to detect lateral gene transfer events are mainly
based on
functional coding DNA characteristics. In this paper, we propose the use of DNA
traits not depending on protein coding requirements. We introduce several
semilocal variables that depend on DNA primary sequence and that reflect
thermodynamic as well as physico-chemical magnitudes that are able to tell apart
the genome of different organisms. After combining these variables in a neural
classificator, we obtain results whose power of resolution go as far as to
detect the exchange of genomic material between bacteria that are
phylogenetically close.

\end{abstract}

\section{Introduction}

There is a general agreement that horizontal gene transfer (HGT) is important in
genome evolution. To which degree is still a matter of debate. The discussion
oscillates between two extreme positions: first, the idea that the rate
of transfer and its impact are of such a magnitude as to be the ``essence of
phylogeny'' -at least for Prokaryotic organisms \cite{Doolittle01} and, on
the other hand, the researchers who opine that the role of HGT in evolution has
been overestimated \cite{Kurland} and that, while a matter of
interest, it is not relevant when compared to other causes of genomic evolution
like paralog duplication and secondary gene looses. Most probably, the real
weight of HGT in evolution is posed somewhere between these extremes.
Independently of the outcome of
the discussion, there is a general agreement that it is relevant to detect
events of HGT.

There are several proposed methods to detect HGT.  They can be classified
into four categories: deviant composition, anomalous phylogenetic distribution,
abnormal sequence similarity and incongruent phylogenetic trees
\cite{Eisen, Ragan, Philippe}.

Deviant composition methods are based on the
different phenotypic characteristics among divergent genomes. They are mainly
focused on bias in GC contents or codon usage \cite{Mrazek} and bias in the
nucleotide
composition in the third and first codon position  \cite{Lawrence}. Deviant
genes
might exist for reasons other than HGT, and only recently transferred
genes
would be detected by this method \cite{Eisen, Hooper}.
Also, this group of methods normally does not detect transferred genes from
phenotypically similar genomes. 

Anomalous phylogenetic distribution is based on the identification of homologous
genes shared by genomes in disjunct phylogenetic lineages and its absence in
close relatives (in one or both lineages). However, polyphyletic gene looses and
rapid sequence divergence can mislead the identification of HGT
\cite{Eisen}.

Abnormal sequence similarity is based on the assumption of overall similarity as
a measure of phylogenetic relatedness. Usually, BLAST searches (or other similar
algorithms) are used to detect sequences in one genome more similar to sequences
in divergent genomes than those sequences found in phylogenetic closer genomes
(the phylogenetic relationships between genomes are set prior the analysis
according to some other criterion like rRNA phylogeny). While these methods work
fast, they are not fully reliable because the similarity between a gene in two
different species can be explained by a number of phenomenons besides HGT. For
instance, evolutionary rate variation can lead to misleading results in the
identification of HGT genes, both as false positives and false negatives
\cite{Eisen}.

Phylogenetic analysis is often considered to be perhaps the best way to
investigate the occurrence of HGT because it remains the only one to reliably
infer historical events from gene sequences \cite{Eisen};
Accordingly, incongruent phylogenetic trees between different families of genes
will be caused by HGT; however conflicting phylogenies can be a result of either
artifacts of phylogenetic reconstruction, HGT or unrecognized paralogy
\cite{Zhaxybayeva}.

In this study, we propose a method that does not depend on DNA functional
traits and due to this reason is no longer correct to say that it detects HGT
because it might also detect transfer of non-coding DNA (ncDNA). From now on, we will refer to
 horizontal genetic material transfer (HGMT). We use eight variables that can be
measured
over a set of windows covering whole genomes and combine them with an Artificial
Neural
Network (NN). The field of using DNA measurables
other than those derived from protein coding requirements have been largely
ignored. Up to our knowledge, there are no methods based exclusively on
structural DNA traits to detect neither HGT or HGMT events.

\section{Methods}

Our approach was to take a pair of DNA sequences --genomes in case of
prokaryotes, chromosomes for eukaryotes-, the first one is the \textit{donor}
genome and the second one the \textit{acceptor}. The data were taken
from Genbank release 24.

To calculate the variables that characterize locally
the DNA sequences, a window is placed over the chromosomes
(Figure~\ref{fig:01}). The window can  slide over the
sequences or can be put randomly (see below).

\begin{figure}[!h]
\centerline{\includegraphics[scale=0.45]{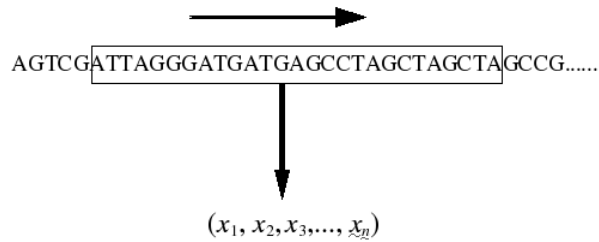}}
\caption{A window is placed over DNA sequence. On every position, some
primary DNA sequence variables $(x_1, x_2,\dots, x_n$) are calculated. For the
NN prediction stage, this window slides along the DNA sequence.
(see text)}\label{fig:01}
\end{figure}

\begin{figure}[!h]
\centerline{\includegraphics[scale=0.37]{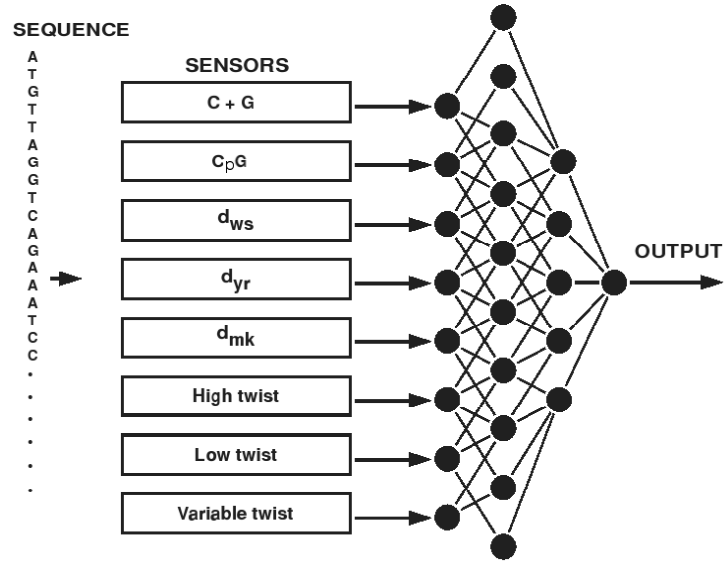}}
\caption{The sensors are mixed in a Multilayer Perceptron. The
output can be zero or one depending on to which class the DNA
sequence belongs.The MLP in the figure does not necessarily have
 the architecture used in the study.}\label{fig:02}
\end{figure}

We used a "classic NN approach" --a backpropragation multilayer
perceptron (MLP), very similar to the model reported by Uberbacher
\cite{Uberbacher} (Figure~\ref{fig:02})

The novelty and main contribution of this paper is the set
of measurables we use to evaluate a DNA sequence. Following the
nomenclature of Uberbacher we will call them {\it sensors}.

For the training stage of the MLP A window of
fixed length (300bp unless otherwise stated) was placed repeatedly over both
genomes
at random independent positions and
every time eight sensors were evaluated over the subsequence in the window and
were used feed and train a MLP with binary output; $'0'$ corresponding to
acceptor DNA sequences and $'1'$ to donor ones. For the prediction stage the
window was allowed to slide along the acceptor sequence and a plot of its
position against the outcome was obtained.

\subsection{Definition of the sensors}

We worked with a total of eight sensors divided in three groups. The first
one includes traditional measures of DNA variance, the second reflects the
DNA local correlations structure and the third one is a measure of the DNA
spatial structure according to the dinucleotide distribution:

\begin{enumerate}

\item This group comprises CG and CpG contents. There is a number of
publicactions reporting the bias of these measures among different organisms.
CpG is well known to tell appart prokaryal and eukaryal lineages
\cite{Shimizu}. Even if the underlying reasons are still unclear \cite{Wang},
it gives a good first clue.

\item In 1995, we  proposed an index of DNA heterogeneity that disclosed
different styles of genomic structural organization \cite{Miramontes}. Given
a DNA sequence, it is translated into the three possible binary sequences using
the groupings purine-pyrimidine ($YR$), weak-strong ($WS$) and amino-keto
($MK$).

the following index is calculated over each binary derivative

\begin{equation}
d=\frac{N_{00}N_{11}-N_{10}N_{01}}{N_0N_1}
\end{equation} 

Where $N_{ij}$ stands for the number of $i$ bases followed by the $j$ base,
where $i$ and $j$ are zero or one. The phenomenology behind this index is
discussed in \cite{Miramontes}.

\item In 1992 the group of R.E. Dickerson \cite{Quintana01} reported the
variability in the DNA structural angles depending on the
dinucleotide steps. Their results can be summarized in the Table~\ref{Tab:01}
(page 345 of the cited reference)

\begin{table}[!h]
%\processtable{This is table caption\label{Tab:01}}
{\begin{tabular}{l|llll}
  & A & C & G & T\\
A & L & I & L & I\\
C & V & L & V & L\\
G & H & H & L & I\\
T & V & H & V & L\\

\end{tabular}}{Matrix of angular variability according to one dimer step along
the DNA sequence}
\end{table}

H stands for high twist profile steps for two stacked consecutive base pairs
along the double helix. They are characterized for having high twist, positive
cup and negative roll angels. The parameters L, I, and V are the Low,
Intermediate and Variable twist.

\end{enumerate}

\section{Results}

\begin{figure}[!h]
\centerline{\includegraphics[scale=0.4]{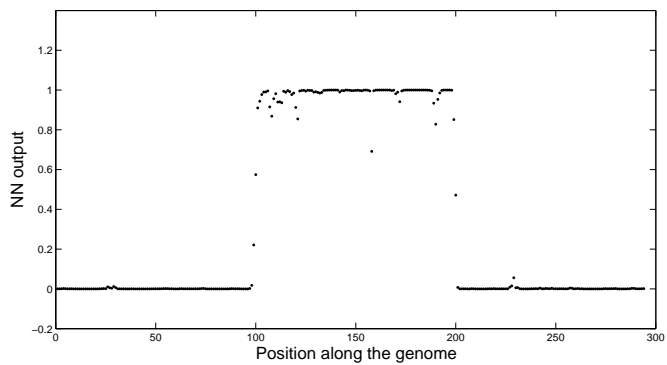}}
\caption{\textit{Mus musculus} chromosome 1 (acceptor) in the horizontal
axis with a sequence of \textit{E. coli} (donor) inserted in the middle. The
ordinate can only be $'0'$ or $'1'$ depending whether the output of the MLP
classify the sequence as an acceptor or as a donor. The horizontal scale
is arbitrary.}\label{fig:03}
\end{figure}

One of the open problems in designing a MLP for pattern detection is to set
the number of neurons in the hidden layer (the number of neurons in the
output layer is determined by the number of classes to classify). In order to
find out the best suited to our interests, we ran several configurations and
tested the resulting output with an artificial problem: To detect a fragment
of \textit{E. coli} (donor) inserted \textit{in silico} in a mouse (\textit{Mus
  musculus}) chromosome (acceptor).  To this end, a set of 20000 fragments of
length 300
of both genomes was picked up randomly to train the MLP. The network
configuration 8-5-1  yielded the results shown in Figure~\ref{fig:03}. With a
300bp sliding window and an overlap of 30bp the
response of the MLP is steadily $'0'$ while the sliding window travel across the
acceptor chromosome and then jumps to $'1'$ when it enters the donor insertion
and goes back to $'0'$ for the rest of the acceptor sequence.

Once the MLP was well tuned, we carried out three case studies:

\begin{enumerate}

\item To detect a prokaryal insertion in a prokaryal genome. In
this case we selected, for no particular reasons,  the genome of
\textit{Archeoglobus fulgidus} as the acceptor genome and \textit{Pseudomonas
aeruginosa} as the donor species in a second {\it in silico } experiment.
Figure~\ref{fig:04} shows that the
results are good enough to encourage one step further.

\begin{figure}[!h]%figure4
\centerline{\includegraphics[scale=0.4]{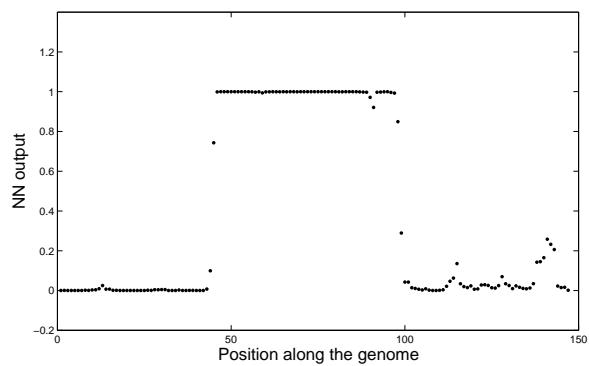}}
\caption{\textit{Archeoglobus fulgidus} (acceptor) in the horizontal
axis with a sequence of \textit{Pseudomonas aeruginosa} (donor) inserted in the
middle. The ordinate can only be $'0'$ or $'1'$ depending whether the output of
the
MLP classify the sequence as an acceptor or as a donor.}\label{fig:04}
\end{figure}

\item Next attempt is the detection of a real horizontal gene transfer already
reported in the literature \cite{Kroll}.  It is a gene (Cu-Zn-superoxide
dismutase) of \textit{Haemophylus ducreyi} inserted in the \textit{Neisseria
mengiditis}
genome.  Figure~\ref{fig:05} shows a clear detection.

\begin{figure}[!h]%figure4
\centerline{\includegraphics[scale=0.4]{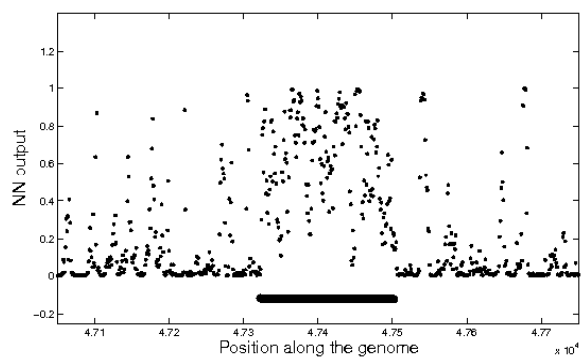}}
\caption{\textit{Neisseria
mengiditis} (acceptor) in the horizontal
axis with a sequence of \textit{Haemophylus ducreyi} (donor) inserted in the
middle. The ordinate can only be $'0'$ or $'1'$ depending whether the output of
the
MLP classify the sequence as an acceptor or as a donor. The
horizontal thick line emphasizes the region of insertion.}\label{fig:05}
\end{figure}

\item Last example is a case of organelle to nuclear genome
transfer. Figure~\ref{fig:06} shows chromosome 2 of {\it Arabidopsis thaliana}
as the acceptor sequence for its own mitochondrion. As the horizontal scale unit
is 30bp, the inserted fragment goes from 107,653 to 116,985 which correspond to
the nucleotides from approximately 3,230,000 to 3,510,000 there is then is a
clear inserted fragment of the order of 270kb which coincides with the
unexpected case of an organelle to nuclear transfer event \cite{Xiaoying}. This
case is important for our claims because approximately more than sixty percent
of the \textit{Arabidopsis thaliana}'s mitochondrial genome is not translated
into aminoacids \cite{Unseld}. Notwithstanding, our method clearly detects the
transfer.

\begin{figure}[!h]%figure4
\centerline{\includegraphics[scale=0.4]{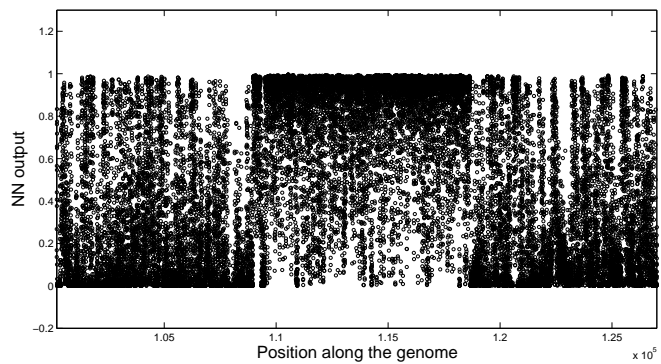}}
\caption{{\it Arabidopsis thaliana}  (acceptor) in the horizontal
axis with a sequence of its own mitochondrial
genome (donor) inserted in the
middle. The horizontal scale unit is 30bp}\label{fig:06}
\end{figure}

\end{enumerate}

\section{Discussion}

In this paper we show the feasibility of using variables not related to the
DNA function as measurables that can be used to detect horizontal exchange of
genetic material between different species. The results are very good and
encourage the further development of this line of research. It is a matter of
future work to build on a complete set of variables with the minimum
size but having the maximum power of resolution.

This paper also contributes to clarify the biology behind the phenomenon
of horizontal gene transfer. For instance,  the degree of
amelioration of horizontally transferred genes is somehow related to the
accuracy to which different methods detects xenologous genes. i,e., the
sequencing of the chromosome 2 from \textit{Arabidopsis thaliana}, has
revealed a large and unexpected organellar-to-nuclear genetic material transfer
event
of the mitochondrial genome (20). The sequence in the nucleus is 99\%
identical to the mitochondrial genome, suggesting that the transfer event 
is very recent. Therefore, amelioration has been negligible, and the
MLP clearly detects the transferred DNA. On the other hand, as shown
in Figure~\ref{fig:05}, the method presented here, effectively identifies DNA
that has been
transferred from \textit{Haemophilus sp}. (a Gamma-proteobacteria) to
\textit{Neisseria meningitidis}, (a member of the Beta-proteobacteria
subdivision) \cite{Kroll}. However, the high degree of spreading of the
points in Figure~\ref{fig:05} suggest that accumulated mutations since
the horizontal gene
transfer event,  have likely ameliorated sodC in \textit{N. meningitidis}. There are 136
nucleotide differences between \textit{H. ducreyi} sodC and the homologous gene
from \textit{N. meningitidis}. If we assume equal rates of substitutions among
the two sequences, then an approximate of 68 new mutations have accumulated
since the horizontal transfer event in each sequences (out of 561 nucleotides in
\textit{N. meningitidis} sodC). This is a substantial amount of change if we
compare to the number of differences among 16S rRNA in the two species (83 differences accumulated in each gene, out of 1544 nucleotides in \textit{N. meningitidis} 16S rRNA) and if we take into account that the transfer event must
have happened after the divergence of the two species. The extent to which the
NN can go in detection when amelioration occurs will be reported elsewhere.

\section*{Acknowledgement}

This research was partially supported by PAPIIT-UNAM grant IN115908.
The investigation is part of Citlali's B.Sc. dissertation in Computer Science at
UNAM.


\begin{thebibliography}{}


\bibitem[Carbone {\it et al}., 2003]{Carbone} Carbone, A., Zinovyev, A. and
K\'ep\`es F. (2003) Codon adaptation index as measure of dominating codon bias.
\textit{Bioinformatics}, \textbf{19}, 2005-2015.

\bibitem[Doolittle, 1999]{Doolittle01} Doolittle, WF. (1999) Phylogenetic
classification and the universal tree. {\it Science}, {\bf 284}, 2124-2129.

\bibitem[Eisen, 2000]{Eisen} Eisen, JA. (2000) Horizontal gene transfer among
microbial genome: new insights form complete genome analysis. {\it Curr Opin
Genet Dev}, {\bf 10}, 606-611.

\bibitem[Hooper and Berg, 2002]{Hooper} Sean D. Hooper and Otto G. Berg (2002)
Detection of genes with atypical nucleotide sequence in microbial genomes. {\it
J Mol Evol} {\bf 54}, 365-375.

\bibitem[Kroll, 1998]{Kroll} Kroll JS, Wilks KE, Farrant JL, Langford PR.
(1998) Natural genetic exchange between Haemophilus and Neisseria: intergeneric
transfer of chromosomal genes between major human pathogens. {\it Proc Natl Acad
Sci USA}, {\it 95}, 12381-5.

\bibitem[Kurland {\it et al}., 2003]{Kurland} Kurland, C.G., Canback, B. And O.
Berg.
(2003). Horizontal gene transfer: A critical view. {\it Proc Natl Acad
Sci USA}, {\bf 100}, 9658-9662.

\bibitem[Lawrence and Ochman, 1997]{Lawrence} Lawrence, JG. and Ochman, H (1997)
Amelioration of bacterial genomes: rates of change and exchange. {\it J Mol
Evol}, {\bf 44}, 383-397.

\bibitem[Mahadevan and Ghosh, 1994]{Mahadevan} Mahadevan I, Ghosh I. (1994)
Analysis of
{\it E.coli} promoter structures using neural networks. {\it Nucleic Acids
Res}, {\bf 22}, 2158-65.

\bibitem[Miramontes {\it et al}., 1995]{Miramontes} Miramontes P, Medrano L,
Cerpa C, Cedergren R, Ferbeyre G and Cocho G.  (1995) Structural and
thermodynamic
properties of DNA uncover different evolutionary histories.  {\it J Mol Evol},
{\bf40}, 698-704.

\bibitem[Mrazek and Karlin, 1999]{Mrazek} Mrazek J, Karlin S. Detecting
alien genes in bacterial genomes (1999). {\it Ann N Y Acad Sci}, {\bf18},
314-29.

\bibitem[Narayanan {\it et al}., 2002]{Narayanan} Narayanan A, Keedwell EC,
Olsson B. (2002) Artificial intelligence techniques for bioinformatics. {\it
Appl Bioinformatics}, {\bf1},1 91-222.

\bibitem[Parbhane {\it et al}., 2000]{Parbhane}  Parbhane RV, Tambe SS, Kulkarni
BD.(2000) ANN modeling of DNA sequences: new strategies using DNA shape code.
{\it Comput Chem}, {\bf24}, 699-711.

\bibitem[Philippe and Douady, 2003]{Philippe} Philippe, H., Douady, CJ.
(2003) Horizontal gene transfer and phylogenetics. {\it Curr Opin
Microbiol}, {\bf 6}, 498-505.

\bibitem[Quintana, 1992]{Quintana01} Quintana, JR., Grzeskowiak,
K., Yanagi, K., Dickerson RE. (1992) Structure of a B-DNA
decamer with a central T-A step: C-G-A-T-T-A-A-T-C-G. {\it J Mol Biol},
{\bf 225}, 379-95.

\bibitem[Ragan, 2001]{Ragan} Ragan, MA. (2001) Detection of lateral gene
transfer among microbial genomes. {\it Curr Opin Genet Dev}, {\bf 11}, 620-626.

\bibitem[Shimizu {\it et al}., 1997]{Shimizu} Shimizu TS, Takahashi K, Tomita M.
(1997) CpG distribution patterns in methylated and non-methylated species. {\it
Gene}, {\bf 205}, 103-7.

\bibitem[Uberbacher {\it et al}., 1991]{Uberbacher}
Uberbacher EC, Mural RJ. (1991) Locating protein-coding regions in human DNA
sequences by a multiple sensor-neural network approach.  {\it Proc Natl Acad Sci
USA},
{\bf 88}, 11261-5.

\bibitem [Unseld {\it et al}., 1997] {Unseld} Unseld, M.,  Marienfeld, RM.,
Brandt P, Brennicke, A. (1997) The mitochondrial genome of Arabidopsis thaliana
contains 57 genes in 366,924 nucleotides. {\it Nat Genet},  {\bf 15},
57-61.

\bibitem[Wang {\it et al}., 2004]{Wang} Wang Y, Rocha EP, Leung FC, Danchin A.
(2004) Cytosine methylation is not the major factor inducing CpG dinucleotide
deficiency in bacterial genomes. {\it J Mol Evol}, {\bf 58}, 692-700.

\bibitem[Wu, 1997]{Wu} Wu CH. (1997) Artificial neural networks for molecular
sequence analysis.  {\it Comput Chem}, {\bf 21}, 237-56.

\bibitem[Xiaoying {\it et al}., 1999]{Xiaoying} Xiaoying, L. {\it et al}.
(1999)  Sequence and analysis of chromosome 2 of the plant \textit{ Arabidopsis
thaliana}. {\it
Nature}, {\bf 402}, 761-768.

\bibitem[Zhaxybayeva {\it et al}., 2004]{Zhaxybayeva} Zhaxybayeva, O., Lapierre,
P., Gogarten, P. (2004) Genome mosaicism and organismal lineages. {\it Trends
Genet}, {\bf20}, 254-260.



\end{thebibliography}
\end{document}